\documentclass[journal]{IEEEtran}
\usepackage{blindtext}
\usepackage{graphicx}
\usepackage{color}
\usepackage{gensymb}

\ifCLASSINFOpdf
\else
\fi
\begin{document}
%
\title{Real-Time Human Detection as an Edge Service Enabled by a Lightweight CNN}


%
%
%

\author{\IEEEauthorblockN{Seyed Yahya Nikouei${^\dagger}$, Yu Chen${^\dagger}$, Sejun Song${^\ddagger}$, Ronghua Xu${^\dagger}$, Baek-Young Choi${^\ddagger}$, Timothy R. Faughnan${^\xi}$\\}
\IEEEauthorblockA{${^\dagger}$Dept. of Electrical and Computing Engineering, Binghamton University, SUNY, Binghamton, NY 13902, USA\\
${^\ddagger}$School of Computing and Engineering, University of Missouri-Kansas City, Kansas City, MO 64110, USA\\
${^\xi}$New York State University Police, Binghamton University, SUNY, Binghamton, NY 13902, USA\\}
 E-mails: \{snikoue1, ychen, rxu22, tfaughn\}@binghamton.edu, \{songsej, choiby\}@umkc.edu}
\maketitle

\begin{abstract}

Edge computing allows more computing tasks to take place on the decentralized nodes at the edge of networks. Today many delay sensitive, mission-critical applications can leverage these edge devices to reduce the time delay or even to enable real-time, online decision making thanks to their on-site presence. Human objects detection, behavior recognition and prediction in smart surveillance fall into that category, where a transition of a huge volume of video streaming data can take valuable time and place heavy pressure on communication networks. It is widely recognized that video processing and object detection are computing intensive and too expensive to be handled by resource-limited edge devices. Inspired by the depthwise separable convolution and Single Shot Multi-Box Detector (SSD), a lightweight Convolutional Neural Network (L-CNN) is introduced in this paper. By narrowing down the classifier's searching space to focus on human objects in surveillance video frames, the proposed L-CNN algorithm is able to detect pedestrians with an affordable computation workload to an edge device. A prototype has been implemented on an edge node (Raspberry PI 3) using openCV libraries, and satisfactory performance is achieved using real-world surveillance video streams. The experimental study has validated the design of L-CNN and shown it is a promising approach to computing intensive applications at the edge. 
%
%

\end{abstract}


%
\IEEEpeerreviewmaketitle

\section{Introduction}

Today the world's population is concentrated in urban areas, because of  the convenient lifestyle in bigger cities, and this urbanization is increasing at an unprecedented scale and speed~\cite{nations2014world}. Meanwhile, significant challenges are rising to handle big data collection and dynamic transition and analytics in real-time or near real-time~\cite{mendoza2009video}. A typical surveillance camera with frame rate of 10 Hz wide area motion imagery (WAMI) sequence alone can generate over 250M of data per second (more than 900G per hour). The data can be significantly more substantial by scale for high frame rate and higher resolution videos~\cite{porter2010wide}. It is shown in recent studies that surveillance application creates heavy workload on the communication networks and most of the data transferred is dominantly video.  The delay and dependence on communication networks still hinder running-time interactions, although suggestions for novel delivery protocols with higher efficiency in mind are presented~\cite{zhao2017buffer},~\cite{zheng2017real}. Storage capacity is another problem that needs attention. Although many companies increased their capacity (typically by 50\%) still a huge volume of streaming video is a burden to store. Usually, after weeks, they are removed to give space for new footage.  

Edge computing extends the realm of information technology (IT) beyond the boundary of cloud computing paradigm by devolving various services (i.e., computation) at the edge of the network. The ubiquitously deployed networked cameras and smart mobile devices are leveraged to have edge computing with promising results to address the challenges in many delay-sensitive, mission-critical applications, such as real-time surveillance. The smart surveillance systems are an ideal fit to the edge-fog-cloud hierarchical architecture. It would be able to significantly improve many tasks, if more intelligence can be migrated to the edge. An example might be fast object detection and tracking. Inspired by the proliferation of machine learning algorithms, a lightweight Convolutional Neural Network (L-CNN) is introduced in this paper. It incurs less pre-processing overhead than other human object detection and classification algorithms. The proposed L-CNN algorithm is carefully tailored to answer the resource constrained environment of an edge device with decent performance. 

In our experimental study, the L-CNN algorithm can process an average of 1.79 and up to 2.06 frames per second (FPS) on the selected edge device, a Raspberry PI 3 board. It meets the design goal considering the limited computing power and the requirement from the application. Intuitively, tracking a car in real-time requires much higher speed than tracking a pedestrian in real-time. Humans tend to walk slowly in comparison to a high frame rate of a camera. Therefore, the real-time human object detection task can be accomplished even if the detection algorithm is executed only twice per second. Notice that this human object detection is meant to detect any human objects newly appear in the frame because other detected objects are tracked with a Tracker algorithm and the objects that walk out are automatically deleted. Based on this rationale, an FPS around two is sufficient for real-time human object detection. 

The major contributions of this work include:

\begin{enumerate}
    \item Customized for real-time application, a lightweight Convolutional Neural Network algorithm is proposed, which enables human objects identification at the edge of the network in a real-time manner;
    \item A prototype of the L-CNN algorithm has been implemented on a Raspberry PI node as the edge device and tested using real-world surveillance videos; and 
    \item An extensive experimental study has been conducted, which compares the L-CNN with the powerful SSD GoogleNet and two other well-known human detection algorithms, Harr-Cascade and HOG + SVM, the effectiveness and correctness of the L-CNN validate that L-CNN is a promising solution for applications like real-time smart surveillance.   
\end{enumerate}


The rest of this paper is structured as follows. In Section \ref{sec:background} the background of the closely related work is provided. The proposed lightweight CNN algorithm is introduced in Section \ref{sec:L-CNN}. Section \ref{sec:exp} reports the comparison study of the L-CNN and other algorithms using a Raspberry PI 3 model B device as the selected edge computing node. Finally, Section \ref{sec:conclusions} concludes this paper with some discussions. 

\section{Background Knowledge and Related Work}
\label{sec:background}

\subsection{Smart Surveillance as an Edge Service}

Migrating functions to the network edge devices will minimize communication delays and message overhead. Also, the decision can be made on-site instantly~\cite{chen2017enabling},~\cite{chen2016smart}. Nevertheless, video processing proves to be a huge computational burden for the edge device and the decision-making task should be outsourced to the fog or even cloud nodes for execution. If the fog node is responsible for decision-making, near real-time execution is achieved. In addition, fine tunning of the decision-making algorithm is also a resource hungry task that can be executed based on historical data in cloud or fog nodes. The most important step is the human detection as a miss detection will lead to false alarms.

Based on the assumptions such as the installed cameras often have a fixed angle or distance to the camera, many algorithms for human detection are introduced that perform well. Knowing the environment will help to choose more relevant images and better features to represent the objects of interest such that the classifiers are able to provide more accurate results. However, since training and operation are vary, they might not be the best choice under different circumstances of lightning conditions, camera angles or background textures. Because of the limitations on computation capabilities and storage constraints ideally, the  human object detection should be as light as possible~\cite{xu2018real}.

\subsection{Convolutional Neural Network}

Due to the occlusion, illumination, and inconstant appearance, the task of human object detection may be more challenging than identifying a moving car or other pattern follower objects in the video frames. However, for tracking purposes it is vital for the detecting algorithm to give an accurate coordination of each human object in the frame. 

More recent deep learning architectures such as GoogLeNet~\cite{szegedy2015going} and Microsoft ResNet~\cite{he2016deep} can classify up to a thousand objects that humans are among them. The authors could not find any deep learning network specially designed for detecting human objects in mind. However, in case of only human objects detection or classification, the same acceptable performance can be reached with a network architecture with much less same size filters.  

Lightweight solutions are studied in recent years, aiming at faster performance with similar accuracy comparing to well-known deep learning networks. SqueezeNet can be considered as one example of this type of networks. The main idea as implied by the name, is to squeeze the network to take less memory while having the same performance as AlexNet~\cite{iandola2016squeezenet}. This is achieved by a fire module in SqueezeNet which combines two sets of filters to reduce computational complexity. SqueezeNet begins with one layer of regular convolution and eight layers of fire module follows that. Finally, another regular convolutional layer comes before the classification network. Along using fewer filters, Iandola et.al., propose to break the computational paradigm and that is why authors use fire module based convolution instead of regular model.

The recently released MobileNet works on resource constrained devices~\cite{howard2017mobilenets}. It presents not only high memory efficiency, but also a very fast forward path. This is reached by a mathematically proven, convolutional model less computational burden while having fewer parameters. MobileNet produces results that are comparable to GoogleNet and ResNet~\cite{he2016deep},~\cite{he2016identity}. The basic idea of the MobileNet is separating regular convolution into two parts, the Depthwise convolutional layer and the Pointwise convolutional layer, to reduce the complexity and accelerate the speed. Due to its high performance, the MobileNet is selected as the base on which our L-CNN algorithm is developed.

\section{Lightweight CNN}
\label{sec:L-CNN}

Edge devices cannot afford the storage space for parameters and weight values of filters in regular deep neural networks. Even after the time consuming and computing intensive training are outsourced to the cloud or other powerful nodes, they still need a huge volume of RAM. For example, the most well-known ResNet has three architectures with 51, 101, or 201 layers respectively, which implies a huge number of parameters to be loaded into small devices' memory. 

In this work, the L-CNN is proposed to fit in the edge computing devices. The first step of the L-CNN is employing the Depthwise Separable Convolution ~\cite{sifrerigid} to reduce the computational cost of the CNN itself, without much sacrificing the accuracy of the whole network. Also, the network is specialized for human detection to reduce the unnecessary huge filter numbers in each layer. Secondly, because the main goal here is to have the surveillance system that can protect humans, the light weighted architecture with the best performance is chosen. The accurate position of the pedestrian is needed to make decisions based on their movement. Thus, state of the art SSD architecture is fused to CNN architecture have fast and reliable results for object detection.

\subsection{Depthwise Separable Convolution}
\label{sec:depth}

By splitting each regular convolution layer into two parts, depthwise separable convolution and pointwise separable convolution~\cite{sifrerigid}, it makes computational complexity more suitable for edge devices. The foundation of the L-CNN is based on these architectures.

Let's begin with a conventional convolution that will receive an input as $F$, with dimensionality of $D_{f}\times D_{f}$ and also has $M$ channels. It will map the input to $G$ as the output, which has $N$ channels and has dimensionality of $D_{g} \times D_{g}$. Filter $K$does this job. As calculated in Eq. \ref{eq:comp1} for a set of $N$ filters, where each of them is $D_{k} \times D_{k}$ and there are $M$ channels considered for each the Computational Complexity can be shown as:


\begin{equation}
 CC_{Conventional}=D_{k} \times D_{k} \times M \times N \times D_{f} \times D_{f}
 \label{eq:comp1}
\end{equation}

\begin{figure}[t]
    \centering
        \includegraphics[width=0.2\textwidth]{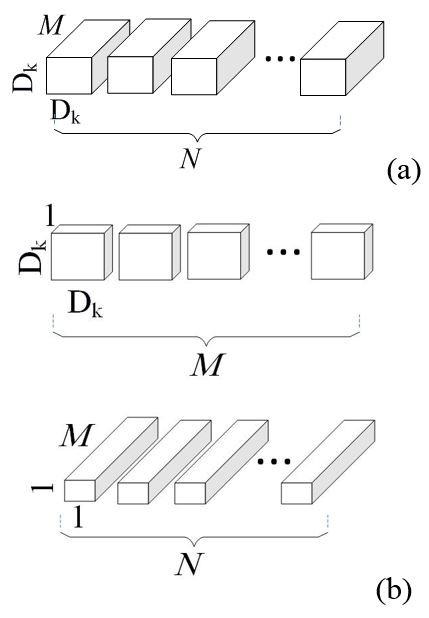}
    \caption{The conventional convolution and Depthwise separable convolution.}
    \label{fig:depth}
    \vspace{-10pt}
\end{figure}

Figure \ref{fig:depth} compares the depthwise separable convolution and the regular convolution. As shown in the figure, the depthwise separable convolution contains two pieces. First part is $M$ channels of $D_{k} \times D_{k} \times 1$ filters that will result in $M$ outputs. This is called a depthwise convolution layer. In second part, filters are of size  $N$ channels of $1 \times 1$ filters and it is called a pointwise convolution layer.


\noindent{where $\hat{K}$ is considered as a depthwise convolutional filter, which has dimension of $D_{k} \times D_{k} \times M$ and the $m_{th}$ filter in $\hat{K}$ is going to be applied on $m_{th}$ $F$. Thus, the Computational Complexity of this approach is:}

\begin{equation}
 CC_{Depth}=D_{k} \times D_{k} \times M \times D_{f} \times D_{f} + N \times M \times D_{f} \times D_{f}
 \label{eq:comp2}
\end{equation}

Reduction of calculation complexity based on Eq. \ref{eq:comp2} and Eq. \ref{eq:comp1}, is  by a factor calculated in Eq. \ref{eq:comp3}~\cite{howard2017mobilenets}. It proves that this approach can be beneficial if used in edge device.

\begin{equation}
 \frac{CC_{Conventional}}{CC_{Depth}} = \frac{1}{N} + \frac{1}{{D_{k}}^2}
 \label{eq:comp3}
\end{equation}


\subsection{The L-CNN Architecture}
\label{sec:human}

\begin{figure}[t]
    \centering
        \includegraphics[width=0.32\textwidth]{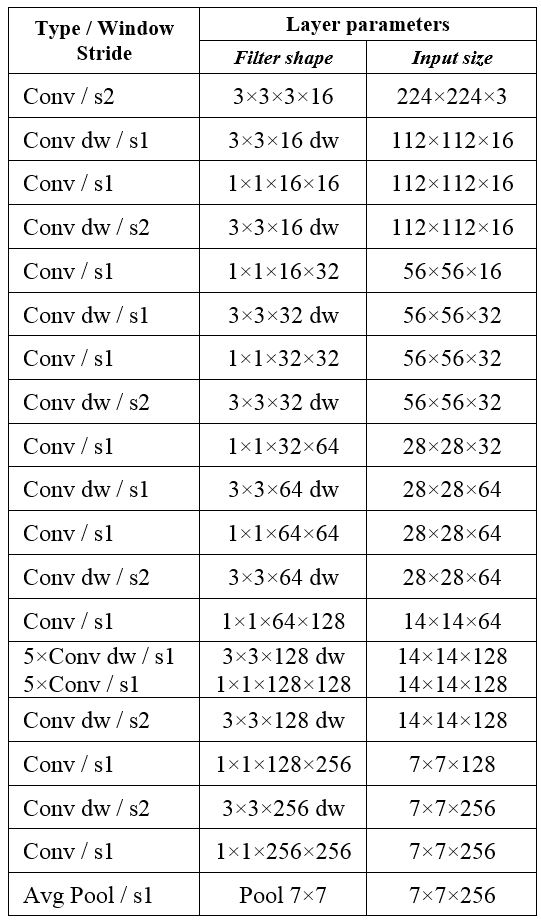}
    \caption{L-CNN network layers specification.}
    \label{fig:L-CNN}
    \vspace{-10pt}
\end{figure}

The proposed L-CNN network architecture has 26 layers considering depthwise and pointwise convolutions as separate layers, which does not count the final classifier, softmax, regression layers to give a bounding box around the detected object. A simple fully connected neural network classifier takes the prior probabilities of each window of objects, identifies the objects within the proposed window, and adds the label for output bounding boxes at the end of the network. Figure \ref{fig:L-CNN} depicts the network filter specifications for each layer.

In this architecture, the pooling layer is added only one time in the final step of the convolutional layers, and downsizing happens with the help of striding parameter in filters. In the first convolutional layer of the L-CNN architecture a conventional convolution takes place, but in the rest of the network depthwise along with pointwise convolutions are used. 

\subsection{Human Object Detection Methods}
\label{sec:R-CNN}

Region-CNN (R-CNN) receives an input frame and using feature maps created by convolutional layers makes proposals of where there is a possibility of existing an object~\cite{girshick2014rich}. This means that R-CNN runs the classifier based on each proposal and tests the probability of an object being present and if the probability is higher than a threshold, the proposal is going to be labeled and so R-CNN treats the network as a couple of separate parts. Using feature maps created, it extracts the bounding boxes and identifies the class of highest probability of objects corresponding to each bounding box. Another model based on R-CNN is the faster R-CNN that reduces the number of proposals and takes other measures to reduce detection time based on real-time application needs. The faster R-CNN performs about 200 times better than the original R-CNN~\cite{ren2017faster}. However, this number depends on the CNN network used. Another method for super fast calculation is YOLO \cite{redmon2016you} which has lower accuracy and was not considered because of this flaw. 

A method faster than R-CNN is the Single Shot Multi-Object Detector (SSD)~\cite{liu2016ssd} which was introduced in late 2016. Based on a unified framework for detection of an object, SSD uses a single network. Hence it will not take the convolutional layers output and run classifier multiple times, it creates a single network that detects objects. SSD modifies the proposal generator to get class probability instead of the existence of an object in the proposal and also instead of having the classical sliding window and checking in each window for an object to report, in the beginning layers of convolutional filters, it will create a set of default bounding boxes over different aspect ratios. Then, with each feature map through convolutional layers, they will be scaled along the image itself. In the end, it will check for each object category presence based on the prior probabilities of the objects in the bounding box. Finally, it adjusts the bounding box to better fit the detection with linear regression~\cite{liu2016ssd}. The proposed L-CNN network uses feature maps from layers 17, 15, 13, 11 to create prior probability map.

\section{Experimental Results}
\label{sec:exp} 

\subsection{Experimental Setup}

In this study, a Raspberry PI 3 Model B with ARMv7 1.2 GHz processor and 1 GB of RAM is used as an edge device of choice. Memory profiler is an application used for recording memory utilization when executing algorithms where SSD-GoogleNet and L-CNN have the same training databases. Averaging in 30 seconds Table \ref{table:fps} shows FPS performance of different algorithms. Other than the L-CNN in Table \ref{table:fps}, other CNNs need to be SSD to be compared here.

\subsection{Results and Discussions}

\begin{table}[ht] \centering{
\begin{tabular}{|l|l|l|l|l|}
  \hline
  \textbf{Algorithms} & \textbf{FPS} & \textbf{CPU(\%)} & \textbf{Memory(MB)} & \textbf{FPR(\%)}\\ \hline
  Haar Cascaded & 1.82  & 76.9 & 111.6 & 26.3 \\ \hline
  HOG + SVM & 0.30 & 93.0 & 139.3 & 14.5 \\ \hline
  SSD GoogleNet & 0.39 & 84.7 & 320.4 & 5.3 \\ \hline
  L-CNN & 1.79  & 75.7 & 122.5 & 6.6  \\ \hline
\end{tabular}
\vspace{5pt}
\caption{Performance in FPS, CPU, Memory Utility and Average False Positive Rate (FPR\%).}
\vspace{-10pt}
\label{table:fps}
}
\end{table}

In Table \ref{table:fps}, the fastest algorithm is the Haar Cascaded, the proposed L-CNN is the second and very close to the best. Table \ref{table:fps} also shows that Haar Cascaded is the best in terms of resource efficiency, and again the L-CNN is the second and very close. However, in terms of average false positive rate (FPR) L-CNN achieved a very decent performance (6.6\%) that is much better than that of Harr Cascaded (26.3\%). In fact, the L-CNN's accuracy is comparable with SSD GoogleNet (5.3\%), but the later features a much higher resource demanding and an extremely low speed (0.39 FPS). In contrast, the average speed of L-CNN is 1.79 FPS and a peak performance of 2.06 FPS is obtained. It is the highest and more than six times faster than the HOG+SVM algorithm (0.3 FPS), which is computationally heavy. Also, in Table \ref{table:fps} the memory used by different algorithms are shown  as an average value in 30 seconds of run time. The GoogleNet needs more memory due to their huge number of filters and their architecture that needs to be loaded in memory at runtime. The reason that GoogleNet is not taking a huge memory in contrary to other reports is that this is an SSD based GoogleNet and images that are used for training have only 21 classes instead of usual one thousand classes. Obviously, with fewer classes, less memory is needed. 

In calculation of these accuracy measures, real-life surveillance video is used. Because there is only one usage model, accuracy reported here may be higher than general purpose usage reported in other literatures.


\begin{figure}[t]
        \centering
        \includegraphics[width=0.38\textwidth]{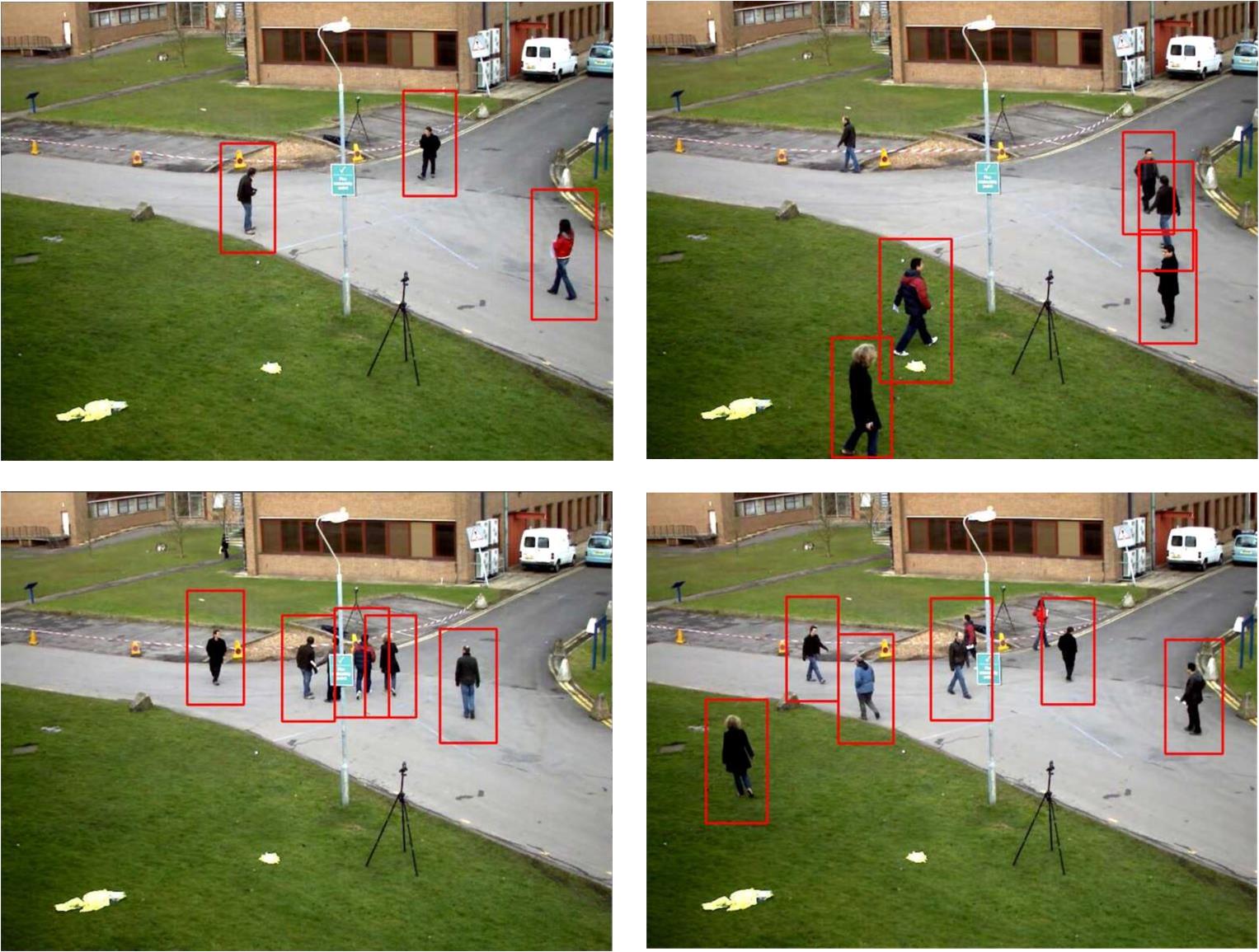}
        \caption{SSD-GoogleNet.}
        \label{fig:r_GoogleNet}
\end{figure}

\begin{figure}[t]
        \centering
        \includegraphics[width=0.38\textwidth]{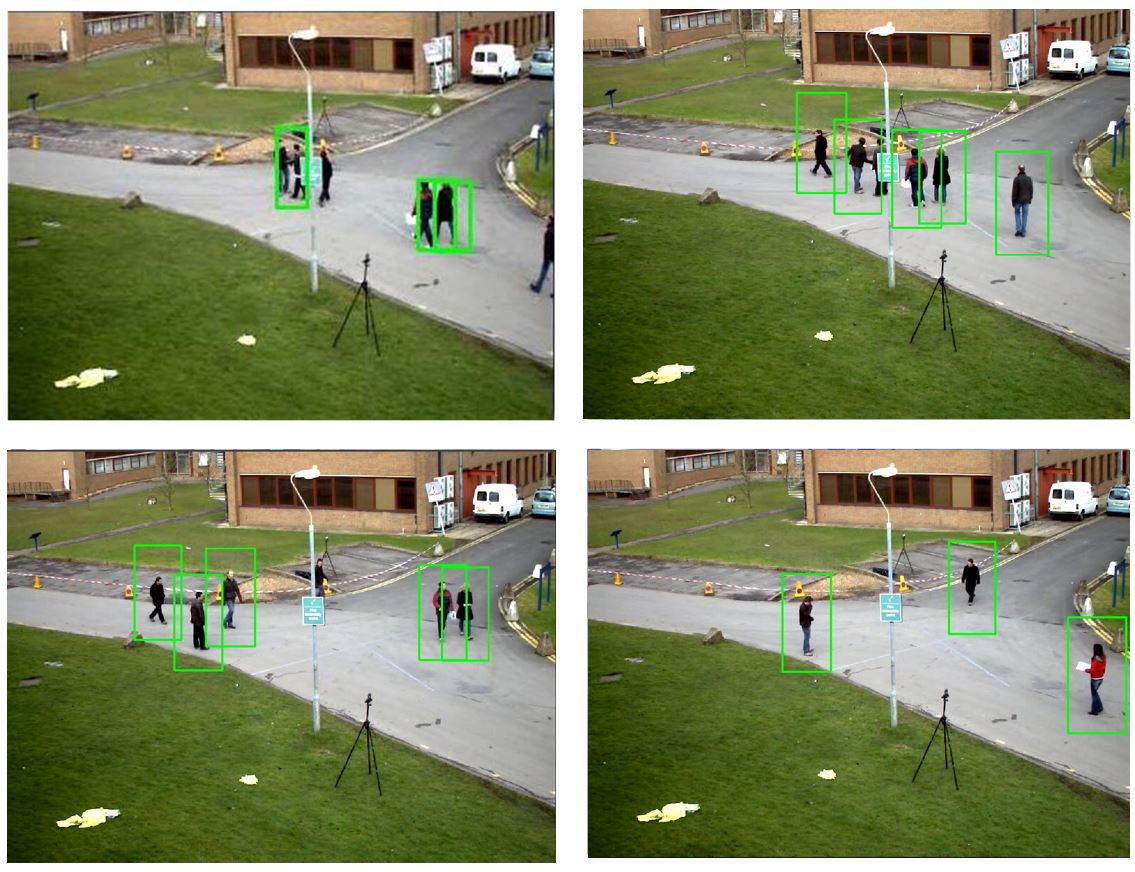}
        \caption{L-CNN.}
        \label{fig:r_L-CNN}
        \vspace{-10pt}
\end{figure}

Figures \ref{fig:r_GoogleNet} and \ref{fig:r_L-CNN} are the results of GoogleNet and L-CNN in processing a sample surveillance video. The video is resized as the input for the algorithms to $224 \times 224$ pixels and because the smaller image size is, the less computation resource requires, they can be compared to each other. It is worth mentioning that figures included in this paper are footage from an online open source video because surveillance videos are not allowed to be exposed to the public.

The high accuracy achieved by the CNNs are well presented in Figures \ref{fig:r_GoogleNet} and \ref{fig:r_L-CNN}. GoogleNet can detect almost all human objects with very low false positive rate of 5.3\%, where performance of the proposed L-CNN is very close. It should be highlighted that while the L-CNN takes three times less memory, its performance is not dropped significantly. In addition, because of fewer number of filters in each layer, the L-CNN achieved a much faster speed than GoogleNet did.

\begin{figure}[t]
        \centering
        \includegraphics[width=0.38\textwidth]{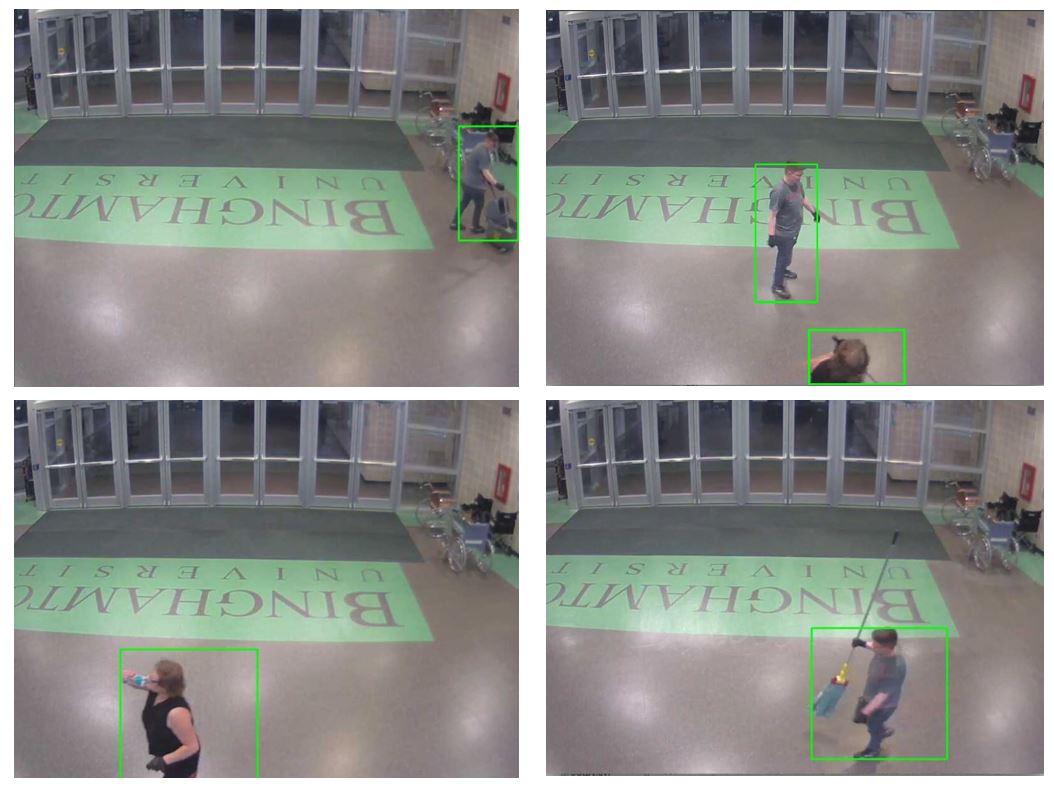}
        \caption{L-CNN: human object from variant angles and distances.}
        \label{fig:r_L-CNN2}
\end{figure}

Figure \ref{fig:r_L-CNN2} focuses on the results of the L-CNN algorithm in case when the human objects of interest appear from different angles. It is harder for classifiers, because sometimes, as seen in Fig. \ref{fig:r_L-CNN2}, the human body is not shown completely in the frame. In the left-bottom subfigure, both two legs of the pedestrian are not visible. Many algorithms such as HOG+SVM either cannot identify human body presence, or by changing parameters very high false positive rate is incurred along with these type of detections. These experimental results verified that the L-CNN is very promising in handling these difficult situations, which were often undetectable.

Table \ref{table:cnns} compares different CNN architectures with the L-CNN algorithm. This test is performed on a desktop machine so that bigger networks such as VGG that is not an SSD object detection architecture can be loaded. It requires up to 20 times more memory space and justifies our presumption that huge networks are not suitable in edge devices. it should be noted that SqueezeNet is not an SSD compatible network, too. The advantage of L-CNN then can be seen in its faster and more accurate performance against MobileNet and SqueezeNet respectively.

\begin{table}[ht] \centering{
\begin{tabular}{|l|l|}
  \hline
  \textbf{Algorithms} & \textbf{Memory (MB)} \\ \hline
  VGG & 2459.8  \\ \hline
  SSD GoogleNet & 320.4   \\ \hline
  SqueezeNet & 145.3  \\ \hline
  MobileNet & 172.2  \\ \hline
  L-CNN & 139.5    \\ \hline
\end{tabular}
\vspace{5pt}
\caption{Memory utility of CNNs.}
\vspace{-10pt}
\label{table:cnns}
}
\end{table}

\section{Conclusions}
\label{sec:conclusions}

In this paper, leveraging the Depthwise Separable Convolutional network, a lightweight CNN is introduced for human object detection at the edge. This model was trained using parts of ImageNet and VOC07 datasets which contains the coordination of the objects of interest.  In the experimental test, L-CNN achieves a Maximum 2.03 and an Average 1.79 FPS, which is comparable with very fast Haar Cascaded algorithm and also has false positive rate of 6.6\% . This verifies that the L-CNN is promising for smart surveillance tasks with decent accuracy and reasonable processing speed. Furthermore, through some more challenging experimental scenarios, it is shown that the L-CNN can handle complex situations where human objects are not completely in the frame.

\ifCLASSOPTIONcaptionsoff
  \newpage
\fi



%

\bibliographystyle{IEEEtranS}

\bibliography{./L-CNN}

\begin{thebibliography}{10}
\providecommand{\url}[1]{#1}
\csname url@samestyle\endcsname
\providecommand{\newblock}{\relax}
\providecommand{\bibinfo}[2]{#2}
\providecommand{\BIBentrySTDinterwordspacing}{\spaceskip=0pt\relax}
\providecommand{\BIBentryALTinterwordstretchfactor}{4}
\providecommand{\BIBentryALTinterwordspacing}{\spaceskip=\fontdimen2\font plus
\BIBentryALTinterwordstretchfactor\fontdimen3\font minus
  \fontdimen4\font\relax}
\providecommand{\BIBforeignlanguage}[2]{{%
\expandafter\ifx\csname l@#1\endcsname\relax
\typeout{** WARNING: IEEEtranS.bst: No hyphenation pattern has been}%
\typeout{** loaded for the language `#1'. Using the pattern for}%
\typeout{** the default language instead.}%
\else
\language=\csname l@#1\endcsname
\fi
#2}}
\providecommand{\BIBdecl}{\relax}
\BIBdecl

\bibitem{chen2017enabling}
N.~Chen, Y.~Chen, E.~Blasch, H.~Ling, Y.~You, and X.~Ye, ``Enabling smart urban
  surveillance at the edge,'' in \emph{2017 IEEE International Conference on
  Smart Cloud (SmartCloud)}.\hskip 1em plus 0.5em minus 0.4em\relax IEEE, 2017,
  pp. 109--119.

\bibitem{chen2016smart}
N.~Chen, Y.~Chen, S.~Song, C.-T. Huang, and X.~Ye, ``Smart urban surveillance
  using fog computing,'' in \emph{Edge Computing (SEC), IEEE/ACM Symposium
  on}.\hskip 1em plus 0.5em minus 0.4em\relax IEEE, 2016, pp. 95--96.

\bibitem{girshick2014rich}
R.~Girshick, J.~Donahue, T.~Darrell, and J.~Malik, ``Rich feature hierarchies
  for accurate object detection and semantic segmentation,'' in
  \emph{Proceedings of the IEEE conference on computer vision and pattern
  recognition}, 2014, pp. 580--587.

\bibitem{he2016deep}
K.~He, X.~Zhang, S.~Ren, and J.~Sun, ``Deep residual learning for image
  recognition,'' in \emph{Proceedings of the IEEE conference on computer vision
  and pattern recognition}, 2016, pp. 770--778.

\bibitem{he2016identity}
------, ``Identity mappings in deep residual networks,'' in \emph{European
  Conference on Computer Vision}.\hskip 1em plus 0.5em minus 0.4em\relax
  Springer, 2016, pp. 630--645.

\bibitem{howard2017mobilenets}
A.~G. Howard, M.~Zhu, B.~Chen, D.~Kalenichenko, W.~Wang, T.~Weyand,
  M.~Andreetto, and H.~Adam, ``Mobilenets: Efficient convolutional neural
  networks for mobile vision applications,'' \emph{arXiv preprint
  arXiv:1704.04861}, 2017.

\bibitem{iandola2016squeezenet}
F.~N. Iandola, S.~Han, M.~W. Moskewicz, K.~Ashraf, W.~J. Dally, and K.~Keutzer,
  ``Squeezenet: Alexnet-level accuracy with 50x fewer parameters and< 0.5 mb
  model size,'' \emph{arXiv preprint arXiv:1602.07360}, 2016.

\bibitem{liu2016ssd}
W.~Liu, D.~Anguelov, D.~Erhan, C.~Szegedy, S.~Reed, C.-Y. Fu, and A.~C. Berg,
  ``Ssd: Single shot multibox detector,'' in \emph{European conference on
  computer vision}.\hskip 1em plus 0.5em minus 0.4em\relax Springer, 2016, pp.
  21--37.

\bibitem{mendoza2009video}
O.~Mendoza-Schrock, J.~A. Patrick, and E.~P. Blasch, ``Video image registration
  evaluation for a layered sensing environment,'' in \emph{Aerospace \&
  Electronics Conference (NAECON), Proceedings of the IEEE 2009
  National}.\hskip 1em plus 0.5em minus 0.4em\relax IEEE, 2009, pp. 223--230.

\bibitem{nations2014world}
U.~Nations, ``World urbanization prospects: The 2014 revision, highlights.
  department of economic and social affairs,'' \emph{Population Division,
  United Nations}, 2014.

\bibitem{porter2010wide}
R.~Porter, A.~M. Fraser, and D.~Hush, ``Wide-area motion imagery,'' \emph{IEEE
  Signal Processing Magazine}, vol.~27, no.~5, pp. 56--65, 2010.

\bibitem{redmon2016you}
J.~Redmon, S.~Divvala, R.~Girshick, and A.~Farhadi, ``You only look once:
  Unified, real-time object detection,'' in \emph{Proceedings of the IEEE
  conference on computer vision and pattern recognition}, 2016, pp. 779--788.

\bibitem{ren2017faster}
S.~Ren, K.~He, R.~Girshick, and J.~Sun, ``Faster r-cnn: Towards real-time
  object detection with region proposal networks,'' \emph{IEEE transactions on
  pattern analysis and machine intelligence}, vol.~39, no.~6, pp. 1137--1149,
  2017.

\bibitem{sifrerigid}
L.~Sifre, ``Rigid-motion scattering for image classification, 2014,'' Ph.D.
  dissertation, Ph. D. thesis, PSU.

\bibitem{szegedy2015going}
C.~Szegedy, W.~Liu, Y.~Jia, P.~Sermanet, S.~Reed, D.~Anguelov, D.~Erhan,
  V.~Vanhoucke, and A.~Rabinovich, ``Going deeper with convolutions,'' in
  \emph{Proceedings of the IEEE conference on computer vision and pattern
  recognition}, 2015, pp. 1--9.

\bibitem{xu2018real}
R.~Xu, S.~Y. Nikouei, Y.~Chen, S.~Song, A.~Polunchenko, C.~Deng, and
  T.~Faughnan, ``Real-time human object tracking for smart surveillance at the
  edge,'' in \emph{the IEEE International Conference on Communications (ICC),
  Selected Areas in Communications Symposium Smart Cities Track}.\hskip 1em
  plus 0.5em minus 0.4em\relax IEEE, 2018.

\bibitem{zhao2017buffer}
P.~Zhao, W.~Yu, X.~Yang, D.~Meng, and L.~Wang, ``Buffer data-driven adaptation
  of mobile video streaming over heterogeneous wireless networks,'' \emph{IEEE
  Internet of Things Journal}, 2017.

\bibitem{zheng2017real}
X.~Zheng and Z.~Cai, ``Real-time big data delivery in wireless networks: a case
  study on video delivery,'' \emph{IEEE Transactions on Industrial
  Informatics}, 2017.

\end{thebibliography}

%





\end{document}